\documentclass[10pt,twocolumn,letterpaper]{article}

\usepackage[dvipsnames]{xcolor}  

\usepackage{cvpr}
\usepackage{times}
\usepackage{epsfig}
\usepackage{graphicx}
\usepackage{amsmath}
\usepackage{amssymb}
\usepackage{booktabs}
\usepackage{tabularx}

\usepackage{mystyle}
\usepackage{multirow}
\usepackage{algorithmic}
\usepackage{algorithm}
\usepackage{resizegather}
\usepackage{makecell}
\usepackage{afterpage}

\graphicspath{{./images/}}

\renewcommand{\paragraph}[1]{\vspace{5pt}\textbf{#1}\,}

\usepackage[pagebackref=true,breaklinks=true,letterpaper=true,colorlinks,bookmarks=false]{hyperref}

\cvprfinalcopy 

\def\cvprPaperID{9308}

\begin{document}

\title{Factorized Higher-Order CNNs\\ with an Application to Spatio-Temporal Emotion Estimation}

\author{Jean Kossaifi\thanks{Joint first authors.}\, \thanks{Jean Kossaifi, Yannis Panagakis and Maja Pantic were with Samsung AI Center, Cambridge and Imperial College London.}\\\vspace{-2pt}
\small NVIDIA \\
\and Antoine Toisoul\footnotemark[1] \\
\small Samsung AI Center
\and Adrian Bulat\\
\small Samsung AI Center
\and Yannis Panagakis\footnotemark[2] \\ \vspace{-2pt}
\small University of Athens
\and Timothy Hospedales\\
\small Samsung AI Center
\and Maja Pantic\footnotemark[2] \\
\small Imperial College London\vspace{10pt}
}

\maketitle

\begin{abstract}

Training deep neural networks with 
spatio-temporal (i.e., 3D) or multidimensional convolutions of higher-order is computationally challenging
due to millions of unknown parameters across dozens of layers. To alleviate this, one approach is to apply low-rank tensor decompositions to convolution kernels in order to compress the network and reduce its number of parameters.
Alternatively, new convolutional blocks, such as MobileNet, can be directly designed for efficiency.
In this paper,  we unify these two approaches by proposing a tensor factorization framework
for efficient multidimensional (separable) convolutions of higher-order. Interestingly, the proposed framework
enables a novel higher-order transduction, allowing to train a network on a given domain (e.g., 2D images or N-dimensional data in general) and 
using transduction to generalize to higher-order data such as videos (or (N+K)--dimensional data in general),
capturing for instance temporal dynamics while
preserving the learnt spatial information.

We apply the proposed methodology, coined CP-Higher-Order Convolution (\emph{HO-CPConv}), to spatio-temporal facial emotion analysis. Most existing facial affect models focus on static imagery and discard all temporal information.
This is due to the above-mentioned burden of training 3D convolutional nets and the lack of large bodies of video data annotated by experts. We address both issues with our proposed framework. Initial training is first done on static imagery before using transduction to generalize to the temporal domain. We demonstrate superior performance on three challenging large scale affect estimation datasets, AffectNet, SEWA, and AFEW-VA.

\end{abstract}

\section{Introduction}

With the unprecedented success of deep convolutional neural networks came the quest for training always deeper networks. However, while deeper neural networks give better performance when trained appropriately, that depth also translates into memory and computationally heavy models, typically with tens of millions of parameters. This is especially true when training training higher-order convolutional nets --e.g. third order (3D) on videos. However, such models are necessary to perform predictions in the spatio-temporal domain and are crucial in many applications, including action recognition and emotion recognition.

In this paper, we depart from conventional approaches and propose a novel factorized multidimensional convolutional block that achieves superior performance through efficient use of the structure in the data.
In addition, our model can first be trained on the image domain and extended seamlessly to transfer performance to the temporal domain. This novel transduction method is made possible by the structure of our proposed block. In addition, it allows one to drastically decrease the number of parameters, while improving performance and computational efficiency and it can be applied to already existing spatio-temporal network architectures such as a ResNet3D~\cite{tran2018closer}. Our method leverages a CP tensor decomposition, in order to separately learn the disentangled spatial and temporal information of 3D convolutions. This improves accuracy by a large margin while reducing the number of parameters of spatio-temporal architectures and greatly facilitating training on video data.

\begin{figure*}[ht]
  \centering
  \includegraphics[width=\linewidth]
  {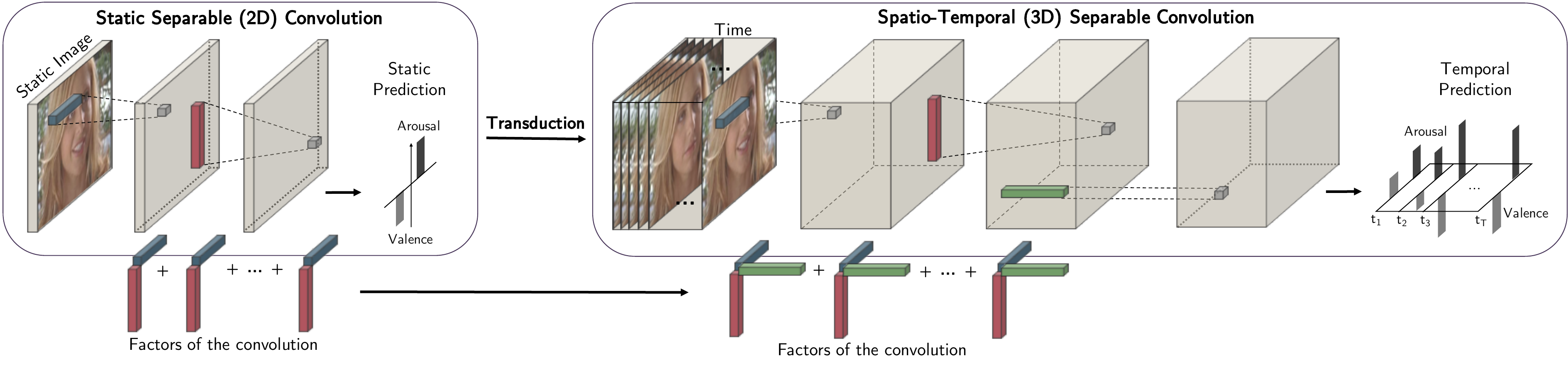}
  \caption{\textbf{Overview of our method}, here represented for a single channel of a single input. We start by training a 2D CNN with our proposed factorized convolutional block on static images \textbf{(left)}. We then apply transduction to extend the model from the static to the spatio-temporal domain \textbf{(right)}. The pretrained spatial factors (blue and red) are first kept fixed, before jointly fine-tuning all the parameters once the temporal factors (green) have been trained.}
  \label{fig:summary}
\end{figure*}

\paragraph{In summary, we make the following contributions:}
\begin{itemize}
  \setlength\itemsep{0em}
    \item We show that many of deep nets architectural improvements, such as \emph{MobileNet} or \emph{ResNet}'s \emph{Bottleneck} blocks, are in fact drawn from the same larger family of tensor decomposition methods~(Section \ref{sec:tensor-convs}). We propose a general framework unifying tensor decompositions and efficient architectures, showing how these efficient architectures can be derived by applying tensor decomposition to convolutional kernels~(Section \ref{ssec:efficient-conv})
    \item Using this framework, we propose factorized higher-order convolutional neural networks, that leverage efficient general multi-dimensional convolutions. These achieve the same performance with a fraction of the parameters and floating point operations~(Section \ref{sec:ho-cp-conv}).
    \item Finally, we propose a novel mechanism called higher-order transduction which can be employed to convert our model, trained in \(N\) dimensions, to \(N+K\) dimensions. 
    \item  We show that our factorized higher-order networks outperform existing works on static affect estimation on the AffectNet, SEWA and AFEW-VA datasets.
    \item Using transduction on the static models, we also demonstrate  state-of-the-art results for continuous facial emotion analysis from video on both SEWA and AFEW-VA datasets.
\end{itemize}

\section{Background and related work}

\paragraph{Multidimensional convolutions} arise in several mathematical models across different fields. They are a cornerstone of Convolutional Neural Networks (CNNs)~\cite{krizhevsky2012imagenet,lecun1998gradient}
enabling them to effectively learn from high-dimensional data by mitigating the curse of dimensionality~\cite{poggio2018theory}.
However, CNNs are computationally demanding, with the cost of convolutions being dominant during both training and inference. As a result, there is an increasing interest in improving the efficiency of multidimensional convolutions.
 
Several efficient implementations of convolutions have been proposed. For instance, 
2D convolution can be efficiently implemented as matrix multiplication by converting the convolution kernel to a Toeplitz matrix. However, this procedure requires replicating the kernel values multiple times across different matrix columns in the Toeplitz matrix, thus increasing the memory requirements.
Implementing convolutions via the im2col approach is also memory intensive due to the space required for building the column matrix. 
These memory requirements may be prohibitive for mobile or embedded devices, hindering the deployment of CNNs in resource-limited platforms.

In general, most existing attempts at efficient convolutions are isolated and there currently is no unified framework to study them. In particular we are interested in two different branches of work, which we review next. Firstly, approaches that leverage tensor methods for efficient convolutions, either to compress or reformulate them for speed. Secondly, approaches that directly formulate efficient neural architecture, e.g., using separable convolutions.

\paragraph{Tensor methods for efficient deep networks}
The properties of tensor methods~\cite{janzamin2019spectral,sidiropoulos2017tensor,tensor_decomposition_kolda} make them a prime choice for deep learning. Beside theoretical study of the properties of deep neural networks~\cite{expressive_deep_tensor}, they have been especially studied in the context of reparametrizing existing layers~\cite{yang2017deep}. One goal of such reparametrization is parameter space savings~\cite{Gusak_2019_ICCV}. \cite{novikov2015tensorizing} for instance proposed to reshape the weight matrix of fully-connected layers into high-order tensors with a Tensor-Train (TT)~\cite{oseledets2011tensor} structure. In a follow-up work~\cite{garipov2016ultimate}, the same strategy is also applied to convolutional layers. Fully connected layers and flattening layers can be removed altogether and replaced with tensor regression layers~\cite{kossaifi2018tensor}. These express outputs through a low-rank multi-linear mapping from a high-order activation tensor to an output tensor of arbitrary order. Parameter space saving can also be obtained, while maintaining multi-linear structure, by applying tensor contraction~\cite{kossaifi2017tensor}. 

Another advantage of tensor reparametrization is computational speed-up. In particular, a tensor decomposition is an efficient way of obtaining separable filters from convolutional kernels. These separable convolutions were proposed in computer vision by~\cite{rigamonti2013learning} in the context of filter banks. 
\cite{jaderberg2014speeding} first applied this concept to deep learning and proposed leveraging redundancies across channels using separable convolutions. 
\cite{astrid2017cp,lebedev2015speeding} proposed to apply CP decomposition directly to the (4--dimensional) kernels of pretrained 2D convolutional layers, making them separable. 
The incurred loss in performance is compensated by fine-tuning. 

Efficient rewriting of convolutions can also be obtained using Tucker decompositions instead of CP to decompose the convolutional layers of a pre-trained network~\cite{yong2016compression}. This allows rewriting the convolution as a \(1 \times 1\) convolution, followed by regular convolution with a smaller kernel and another \(1 \times 1\) convolution. In this case, the spatial dimensions of the convolutional kernel are left untouched and only the input and output channels are compressed. Again, the loss in performance is compensated by fine-tuning the whole network.
Finally, \cite{tai2015convolutional} propose 
to remove redundancy in convolutional layers and express these as the composition of two convolutional layers with less parameters. 
Each 2D filter is approximated by a sum of rank--\(1\) matrices. Thanks to this restricted setting, a closed-form solution can be readily obtained with SVD. Here, we unify the above works and propose factorized higher-order (separable) convolutions.

\paragraph{Efficient neural networks}
While concepts such as separable convolutions have been studied since the early successes of deep learning using tensor decompositions, they have only relatively recently been ``rediscovered'' and proposed as standalone end-to-end trainable efficient neural network architectures. 
The first attempts in the direction of neural network architecture optimization were proposed early in the ground-breaking VGG network~\cite{simonyan2014very} where the large convolutional kernels used in AlexNet~\cite{krizhevsky2009learning} were replaced with a series of smaller ones that have an equivalent receptive field size: i.e. a convolution with a $5\times5$ kernel can be replaced by two consecutive convolutions of size $3\times3$. In parallel, the idea of decomposing larger kernels into a series of smaller ones is explored in the multiple iterations of the Inception block~\cite{szegedy2017inception,szegedy2015going,szegedy2016rethinking} where a convolutional layer with a $7\times7$ kernel is approximated with two $7\times1$ and $1\times7$ kernels. \cite{he2016deep} introduced the so-called \emph{bottleneck} module that reduces the number of channels  on which the convolutional layer with higher kernel size ($3\times3$) operate on by projecting back and forth the features using two convolutional layers with $1\times1$ filters. \cite{xie2017aggregated} expands upon this by replacing the $3\times3$ convolution with a grouped convolutional layer that can further reduce the complexity of the model while increasing representational power at the same time. Recently,~\cite{mobilenet} introduced the \emph{MobileNet} architecture where they proposed to replace the $3\times3$ convolutions with a depth-wise separable module: a depth-wise $3\times3$ convolution (the number of groups is equal to the number of channels) followed by a $1\times1$ convolutional layer that aggregates the information. These type of structures were shown to offer a good balance between the performance offered and the computational cost they incur. \cite{mobilenet-v2} goes one step further and incorporates the idea of using separable convolutions in an \emph{inverted bottleneck} module. The proposed module uses $1\times1$ layers to expand and then contract the channels (hence inverted bottleneck) while using separable convolutions for the $3\times3$ convolutional layer.

\paragraph{Facial affect analysis} is the first step towards better human-computer interactions. Early research focused on detecting discrete emotions such as happiness and sadness, based on the hypothesis that these are \emph{universal}. However, this categorization of human affect is limited and does not cover the wide emotional spectrum displayed by humans on a daily basis.
Psychologists have since moved towards more fine grained dimensional measures of affect~\cite{posner2005circumplex,russel2003}.
The goal is to estimate \emph{continuous} levels of valence --how positive or negative an emotional display is-- and arousal --how exciting or calming is the emotional experience. This task is the subject of most of the recent research in affect estimation~\cite{ringeval2019audio,kollias2019deep,kossaifi2017afew,zafeiriou2017aff} and is the focus of this paper. 

Valence and arousal are dimensional measures of affect that vary in time. It is these changes that are important for accurate human-affect estimation. Capturing temporal dynamics of emotions is therefore crucial and requires video rather than static analysis. However, spatio-temporal models are difficult to train due to their large number of parameters, requiring very large amounts of annotated videos to be trained successfully. Unfortunately, the quality and quantity of available video data and annotation collected in naturalistic conditions is low~\cite{automatic_survey_2015}. 
As a result, most work in this area of affect estimation in-the-wild focuses on affect estimation from static imagery~\cite{kossaifi2017afew,Mollahosseini2017}. Estimation from videos is then done on a frame-by-frame basis. Here, we tackle both issues and train spatio-temporal networks that outperform existing methods for affect estimation.

\section{Convolutions in a tensor framework}\label{sec:tensor-convs}

In this section, we explore the relationship between tensor methods and deep neural networks' convolutional layers. Without loss of generality, we omit the batch size in all the following formulas.

\paragraph{Mathematical background and notation}
We denote \(1\myst\)--order tensors (vectors) as \(\myvector{v}\), \(2\mynd\)--order tensor (matrices) as \(\mymatrix{M}\) and tensors of order \( \geq 3\) as \(\mytensor{T}\).
 We denote a regular convolution of \(\mytensor{X}\) with \(\mymatrix{W}\) as \(\mytensor{X} \myconv_n \mymatrix{W}\). For \(1\)--D convolutions, we write the convolution of a tensor \(\mytensor{X} \in \myR^{I_0, \cdots, I_N}\) with a vector \(\myvector{v} \in \myR^{K}\) along the \(n\myth\)--mode as \(\mytensor{X} \myconv_n \myvector{v}\). In practice, as done in current deep learning frameworks~\cite{pytorch}, we use cross-correlation, which differs from a convolution by a flip of the kernel. This does not impact the results since the weights are learned end-to-end. In other words, \(\left( \mytensor{X} \myconv_n \myvector{v}\right)_{i_0, \cdots, i_N}  = \sum_{k = 1}^K \myvector{v}_k \mytensor{X}_{i_0, \cdots, i_{n-1}, i_n + k, i_{n+1}, \cdots, I_N}\).

\subsection{\(1 \times 1\) convolutions and tensor contraction}
 We show that \(1 \times 1\) convolutions are equivalent to a tensor contraction with the kernel of the convolution along the channels dimension.
Let's consider a \(1\times1\) convolution \(\Phi\), defined by kernel \(\mytensor{W}  \in \myR^{T \times C \times 1 \times 1}\) and applied to an activation tensor \(\mytensor{X} \in \myR^{C \times H \times W}\). We denote the squeezed version of \(\mytensor{W}\) along the first mode as \(\mymatrix{W} \in \myR^{T \times C}\).

The tensor contraction of a tensor \(\mytensor{T} \in \myR^{I_0 \times I_1 \times \cdots \times I_N}\) with matrix \(\mymatrix{M} \in \myR^{J \times I_n}\), along the \(n\myth\)--mode (\(n \in \myrange{0}{N}\)), known as \emph{n--mode product}, is defined as \(\mytensor{P} = \mytensor{T} \times_n \mytensor{M}\), with:
$
    \mytensor{P}_{i_0, \cdots, i_N} = \sum_{k=0}^{I_n} 
    \mytensor{T}_{i_0, \cdots, i_{n-1}, k, i_{n+1}, \cdots, i_N} \mymatrix{M}_{i_n, k}
$

By plugging this 
into the expression of \(\Phi(\mytensor{X})\), we readily observe that the \(1 \times 1\) convolution is equivalent with an n-mode product between \mytensor{X} and the matrix \mymatrix{W}:
\begin{equation}
    \label{eq:n-mode-1x1}\nonumber
    \Phi(\mytensor{X})_{t, y, x} =
    \mytensor{X} \myconv \mytensor{W} = 
    \sum_{k=0}^C  \mytensor{W}_{t, k, y, x} \mytensor{X}_{k, y, x}
    = \mytensor{X} \times_0 \mymatrix{W}
\end{equation}

\subsection{Kruskal convolutions}\label{ssec:cp-conv}
Here we show how separable convolutions can be obtained by applying CP decomposition to the kernel of a regular convolution~\cite{lebedev2015speeding}.
We consider a convolution  defined by its kernel weight tensor \(\mytensor{W} \in \myR^{T \times C \times K_H \times K_W}\), applied on an input of size \(\myR^{C \times H \times W}\). Let \(\mytensor{X} \in \myR^{C \times H \times W}\) be an arbitrary activation tensor. If we define the resulting feature map as \(\mytensor{F} = \mytensor{X} \myconv \mytensor{W}\), we have:
\begin{equation}
    \label{eq:convolution}
    \mytensor{F}_{t, y, x} = 
    \myblue{\sum_{k=1}^C}
    \myred{\sum_{j=1}^H}
    \mygreen{\sum_{i=1}^W}
    \mytensor{W}(t, \myblue{k}, \myred{j}, \mygreen{i})
    \mytensor{X}(\myblue{k}, \myred{j + y}, \mygreen{i + x})
\end{equation}

Assuming a low-rank Kruskal structure on the kernel \(\mytensor{W}\) (which can be readily obtained by applying CP decomposition), we can write:
\begin{equation}
    \label{eq:kruskal-kernel}
    \mytensor{W}_{t, s, j, i} =
    {\color{black}{\sum_{r=0}^{R-1}}}
        \mymatrix{U}^{(T)}_{t, r} 
        \mymatrix{U}^{(C)}_{s, r} 
        \mymatrix{U}^{(H)}_{j, r} 
        \mymatrix{U}^{(W)}_{i, r} 
\end{equation}

By plugging~\ref{eq:kruskal-kernel} into~\ref{eq:convolution} and re-arranging the terms, we get:
\begin{equation}
    \label{eq:separable-convolution}\nonumber
    \mytensor{F}_{t, y, x} = 
    \underbrace{\sum_{r=0}^{R-1} \mymatrix{U}^{(T)}_{t, r}
    \underbrace{\left[\mygreen{\sum_{i=1}^W} \mymatrix{U}^{(W)}_{\mygreen{i}, r}
    \underbrace{\left(\myred{\sum_{j=1}^H} \mymatrix{U}^{(H)}_{\myred{j}, r}
    \underbrace{\left[\myblue{\sum_{k=1}^C}  \mymatrix{U}^{(C)}_{\myblue{k}, r}\mytensor{X}(\myblue{k}, \myred{j + y}, \mygreen{i + x})
     \right]}_{\myblue{1\times1 \text{ conv}}}
     \right)}_{\myred{\text{ depthwise conv}}}
    \right]}_{\mygreen{\text{ depthwise conv}}} 
    }_{1\times 1 \text{ convolution}}
\end{equation}

\begin{figure}[!h]
  \centering
  \includegraphics[width=1\linewidth,trim={0 150 0 150},clip]{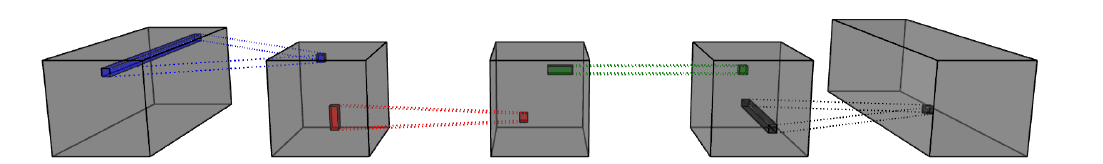}
  \caption{\textbf{Illustration of a 2D Kruskal convolution.}}
  \label{fig:kruskal-conv}
\end{figure}

This allows to replace the original convolution by a series of efficient depthwise separable convolutions~\cite{lebedev2015speeding}, figure~\ref{fig:kruskal-conv}.

\subsection{Tucker convolutions}\label{ssec:tucker-conv}
As previously, we consider the convolution \(\mytensor{F} = \mytensor{X} \myconv \mytensor{W}\). However, instead of a Kruskal structure, we now assume a low-rank Tucker structure on the kernel \(\mytensor{W}\) (which can be readily obtained by applying Tucker decomposition) and yields an efficient formulation~\cite{yong2016compression}. 
We can write:
\begin{equation}
    \label{eq:tucker-kernel}\nonumber
    \mytensor{W}(t, s, j, i) =
    \sum_{r_0=0}^{R_0-1}
    \sum_{r_1=0}^{R_1-1}
    \sum_{r_2=0}^{R_2-1}
    \sum_{r_3=0}^{R_3-1}
        \mytensor{G}_{r_0, r_1, r_2, r_3}
        \mymatrix{U}^{(T)}_{t, r_0} 
        \mymatrix{U}^{(C)}_{s, r_1} 
        \mymatrix{U}^{(H)}_{j, r_2} 
        \mymatrix{U}^{(W)}_{i, r_3}
\end{equation}

Plugging back into a convolution, we get:
\begin{equation}\nonumber
    \mytensor{F}_{t, y, x} = 
    \sum_{k=1}^C
    \sum_{j=1}^H
    \sum_{i=1}^W
    \sum_{r_0=0}^{R_0-1}
    \sum_{r_1=0}^{R_1-1}
    \sum_{r_2=0}^{R_2-1}
    \sum_{r_3=0}^{R_3-1}
        \mytensor{G}_{r_0, r_1, r_2, r_3}
        \mymatrix{U}^{(T)}_{t, r_0} 
        \mymatrix{U}^{(C)}_{k, r_1} 
        \mymatrix{U}^{(H)}_{j, r_2} 
        \mymatrix{U}^{(W)}_{i, r_3}
    \mytensor{X}_{k, j + y, i + x}
\end{equation}

We can further absorb the factors along the spacial dimensions into the core by writing
$ 
    \mytensor{H} = \mytensor{G} 
    \times_2 \mymatrix{U}^{(H)}_{j, r_2}
    \times_3 \mymatrix{U}^{(W)}_{i, r_3}.
$ 

In that case, the expression above simplifies to:
\begin{equation}\label{eq:tucker-conv-basic}
    \mytensor{F}_{t, y, x} = 
    \sum_{k=1}^C
    \sum_{j=1}^H
    \sum_{i=1}^W
    \sum_{r_0=0}^{R_0-1}
    \sum_{r_1=0}^{R_1-1}
        \mytensor{H}_{r_0, r_1, j, i}
        \mymatrix{U}^{(T)}_{t, r_0} 
        \mymatrix{U}^{(C)}_{k, r_1} 
    \mytensor{X}_{k, j + y, i + x}
\end{equation}

In other words, this is equivalence to first transforming the number of channels, then applying a (small) convolution before returning from the rank to the target number of channels. This can be seen by rearranging the terms from Equation~\ref{eq:tucker-conv-basic}:
\begin{gather}\label{eq:tucker-conv}\nonumber
    \mytensor{F}_{t, y, x} =
    \underbrace{
        \mygreen{\sum_{r_0=0}^{R_0-1}}
        \mymatrix{U}^{(T)}_{t, \mygreen{r_0}}
        \left[
            \underbrace{
                \myred{
                    \sum_{j=1}^H
                    \sum_{i=1}^W
                    \sum_{r_1=0}^{R_1-1}
                }
                \mytensor{H}_{\mygreen{r_0}, \myred{r_1}, \myred{j}, \myred{i}}
                \left[
                    \underbrace{
                        \myblue{\sum_{k=1}^C}
                        \mymatrix{U}^{(C)}_{\myblue{k}, \myred{r_1}}
                        \mytensor{X}(\myblue{k}, \myred{j + y}, \myred{i + x})
                    }_{\myblue{1 \times 1 \text{ conv}}}
                \right]
            }_{\myred{H \times W \text{ conv}}}
        \right]
    }_{\mygreen{1 \times 1 \text{ conv}}}
\end{gather}

\begin{figure}[!h]
  \centering
  \includegraphics[width=1\linewidth,trim={0 130 0 150},clip]{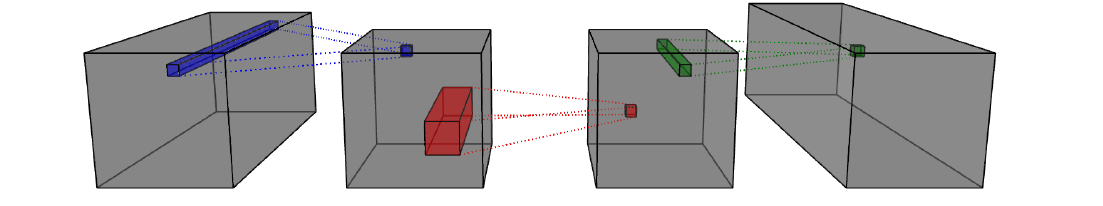}
  \caption{\textbf{Illustration of a Tucker convolution} expressed as a series of small efficient convolutions. Note that this is the approach taken by ResNet for the Bottleneck blocks.}
  \label{fig:tucker-conv}
\end{figure}

In short, this simplifies to the following expression, also illustrated in Figure~\ref{fig:tucker-conv}:
\begin{equation}
    \mytensor{F} = 
    \left(
        \left(\mytensor{X} \times_0 \mymatrix{U}^{(C)} \right)
        \myconv \mytensor{G}
    \right)  \times_0 \mymatrix{U}^{(T)}
\end{equation}

\subsection{Efficient architectures in a tensor framework}\label{ssec:efficient-conv}
While tensor decompositions have been explored in the field of mathematics for decades and in the context of deep learning for years, they are regularly rediscovered and re-introduced in different forms. Here, we revisit popular deep neural network architectures under the lens of tensor factorization. Specifically, we show how these blocks can be obtained from a regular convolution by applying tensor decomposition to its kernel. In practice, batch-normalisation layers and non-linearities are inserted in between the intermediary convolution to facilitate learning from scratch.

\paragraph{ResNet Bottleneck block}
\cite{resnet} introduced a block, coined \emph{Bottleneck block} in their seminal work on deep residual networks. It consists in a series of a \(1 \times 1\) convolution, to reduce the number of channels, a smaller regular (\(3 \times 3\)) convolution, and another \(1 \times 1\) convolution to restore the rank to the desired number of output channels. Based on the equivalence derived in Section~\ref{ssec:tucker-conv}, it is straightforward to see this as applying Tucker decomposition to the kernel of a regular convolution.

\paragraph{ResNext and Xception}
ResNext~\cite{xie2017aggregated} builds on this bottleneck architecture, which, as we have shown, is equivalent to applying Tucker decomposition to the convolutional kernel. In order to reduce the rank further, the output is expressed as a sum of such bottlenecks, with a lower-rank. This can be reformulated efficiently using grouped-convolution~\cite{xie2017aggregated}. In parallel, a similar approach was proposed by~\cite{chollet2017xception}, but without \(1 \times 1\) convolution following the grouped depthwise convolution.

\begin{figure}[!h]
  \centering
  \includegraphics[width=1\linewidth,trim={0 120 0 140},clip]{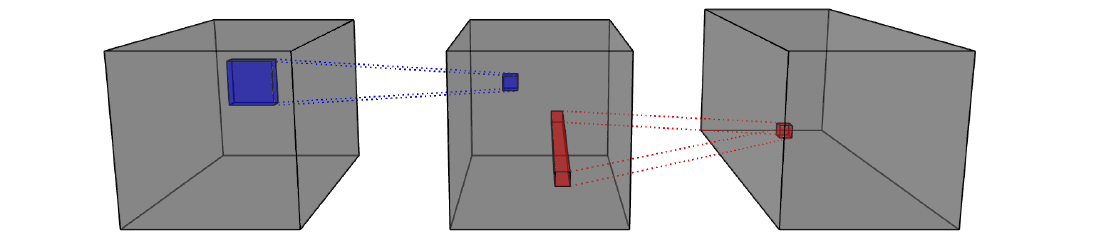}
  \caption{\textbf{MobileNet blocks are a special case of CP convolutions}, without the first convolution, and with spatial factors are combined into one.
  }
  \label{fig:mobilenet-conv}
\end{figure}

\paragraph{MobileNet v1}
MobileNet v1~\cite{mobilenet} uses building blocks made of a depthwise separable convolutions (spatial part of the convolution) followed by a \(1 \times 1\) convolution to adjust the number of output channels. This can be readily obtained from a CP decomposition (Section~\ref{ssec:cp-conv}) as follows: first we write the convolutional weight tensor as detailed in Equation~\ref{eq:kruskal-kernel}, with a rank equal to the number of input channels, i.e. \(R = C\). The first depthwise-separable convolution can be obtained by combining the two spatial \(1\)D convolutions \(\mymatrix{U}^{(H)}\) and \(\mymatrix{U}^{(W)}\). This results into a single spatial factor \(\myblue{\mymatrix{U}^{(S)}} \in \myR^{H \times W \times R}\), such that \(\mymatrix{U}^{(S)}_{j, i, r} = \mymatrix{U}^{(H)}_{j, r} \mymatrix{U}^{(W)}_{i, r} \). 
The \(1 \times 1\) convolution is then given by the matrix-product of the remaining factors, \(\myred{\mymatrix{U}^{(F)}} = \mymatrix{U}^{(T)} \left(\mymatrix{U}^{(C)}\right)\myT \in \myR^{T \times C}\). 
This is illustrated in Figure~\ref{fig:mobilenet-conv}.

\begin{figure}[!h]
  \centering
  \includegraphics[width=1\linewidth,trim={0 130 0 150},clip]{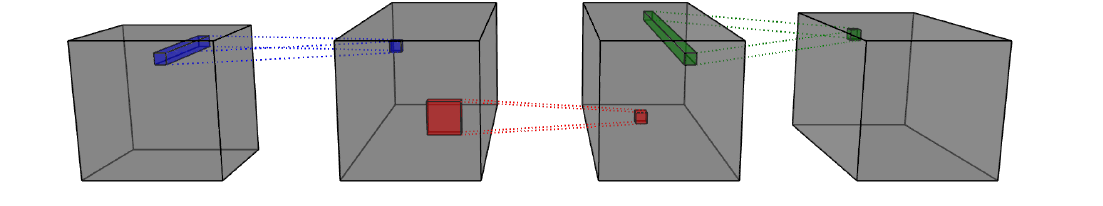}
  \caption{\textbf{MobileNet-v2 blocks are a special case of CP convolutions}, with the spatial factors merged into a depthwise separable convolution.}
  \label{fig:mobilenet-v2-conv}
\end{figure}

\paragraph{MobileNet v2}
MobileNet v2~\cite{mobilenet-v2} employs a similar approach by grouping the spatial factors into one spatial factor \(\mymatrix{U}^{(S)} \in \myR^{H \times W \times R}\), as explained previously for the case of MobileNet. However, the other factors are left untouched. The rank of the decomposition, in this case, corresponds, for each layer, to the expansion factor $\times$ the number of input channels. This results in two \(1 \times 1\) convolutions and a \( 3 \times 3\) depthwise separable convolution. Finally, the kernel weight tensor (displayed graphically in Figure~\ref{fig:mobilenet-v2-conv}) is expressed as:
\begin{equation}
    \label{eq:mobilenet-v2}
    \mytensor{W}_{t, \myblue{s}, \myred{j, i}} =
    {\color{black}{\sum_{\mygreen{r}=0}^{R-1}}}
        \mymatrix{U}^{(T)}_{t, \mygreen{r}} 
        \mymatrix{U}^{(C)}_{\myblue{s}, \mygreen{r}} 
        \mymatrix{U}^{(S)}_{\myred{j, i}, \mygreen{r}} 
\end{equation}
In practice, MobileNet-v2 also includes batch-normalisation layers and non-linearities as well as a skip connection to facilitate learning.

\vspace{-5pt}
\section{Factorized higher-order convolutions}\label{sec:ho-cp-conv}

We propose to generalize the framework introduced above to convolutions of any arbitrary order. Specifically, we express, in the general case, separable ND-convolutions as a series of tensor contractions and 1D convolutions. We show how this is derived from a CP convolution on the N--dimensional kernel. We then detail how to expand our proposed factorized higher-order convolutions, trained in N-dimensions to \((N+1)\) dimensions.

\paragraph{Efficient N-D convolutions via higher-order factorization}
In particular, here, we consider an \(N+1\myth\)--order input activation tensor \(\mytensor{X} \in \myR^{C \times D_0 \times \cdots \times D_{N-1}}\) corresponding to \(N\) dimensions with \(C\) channels.
We define a general, high order separable convolution \(\Phi\) defined by a kernel \(\mytensor{W} \in \myR^{T \times C \times K_0 \times \cdots \times K_{N-1}}\), and expressed as a Kruskal tensor, i.e.
\(\mytensor{W} = \mykruskal{\myvector{\lambda}; \,\,\mymatrix{U}^{(T)}, \mymatrix{U}^{(C)}, \mymatrix{U}^{(K_0)}, \cdots, \mymatrix{U}^{(K_{N-1})}}\).
We can then write:
\vspace{-2pt}
\begin{equation}\nonumber
    \begin{split}
    \Phi (\mytensor{X})_{t, i_0, \cdots, i_{N-1}}  & = 
    \sum_{r = 0}^{R}
    \sum_{s = 0}^C
     \sum_{i_0 = 0}^{K_0}
    \cdots
    \\ 
    &
    \cdots \sum_{i_{N-1} = 0}^{K_{N-1}} \lambda_r \bigl[\mymatrix{U}^{(T)}_{t, r} \mymatrix{U}^{(C)}_{s, r} 
    \mymatrix{U}^{(K_0)}_{i_0, r} 
    \cdots
    \mymatrix{U}^{(K_{N-1})}_{i_{N-1}, r} 
    \mytensor{X}_{s, i_0, \cdots, i_{N-1}} \bigr]
    \end{split}
\end{equation}

By rearranging the terms, this expression can be rewritten as:
\begin{equation}
    \label{eq:ho-conv-2}
    \mytensor{F} = 
    \left(
        \rho \left( \mytensor{X} \times_0 \mymatrix{U}^{(T)} \right)
    \right) \times_0 \left( \mydiag(\myvector{\lambda}) \mymatrix{U}^{(C)} \right)
\end{equation}

where $\rho$ applies the 1D spatial convolutions:
\begin{equation}\nonumber
    \label{eq:rho}
    \rho(\mytensor{X}) = 
    \left(
        \mytensor{X}
        \myconv_1 \mymatrix{U}^{(K_0)}
        \myconv_2 \mymatrix{U}^{(K_1)}
        \myconv \cdots
        \myconv_{N+1} \mymatrix{U}^{(K_{N-1})}
    \right)
\end{equation}

Tensor decompositions (and, in particular, decomposed convolutions) are notoriously hard to train end-to-end~\cite{jaderberg2014speeding,lebedev2015speeding,tai2015convolutional}. As a result, most existing approach rely on first training an uncompressed network, then decomposing the convolutional kernels before replacing the convolution with their efficient rewriting and fine-tuning to recover lost performance. 
However, this approach is not suitable for higher-order convolutions where it might not be practical to train the full N-D convolution.
It is possible to facilitate training from scratch by absorbing the magnitude of the factors into the weight vector $\myvector{\lambda}$. We can also add add non-linearities \(\Psi\) (e.g. batch normalisation combined with RELU), leading to the following expression, resulting in an efficient higher-order CP convolution:
\begin{equation}
    \label{eq:ho-conv-3}
    \mytensor{F} = 
        \rho\left(\Psi
        \left(\mytensor{X} \times_0 \mymatrix{U}^{(T)} \right)
        \right)
    \times_0 \left( \mydiag(\myvector{\lambda}) \mymatrix{U}^{(C)} \right)
\end{equation}

Skip connection can also be added by introducing an additional factor \(\mymatrix{U}^{(S)} \in \myR^{T \times C}\) and using 
\(\mytensor{F}' = \mytensor{X} + \left(\mytensor{F} \times_0 \mymatrix{U}^{(S)} \right) \).

This formulation is significantly more efficient than that of a regular convolution. Let's consider an N-dimensional convolution, with \(C\) input channels and \(T\) output channels, i.e. a weight of size  \(\mytensor{W} \in \myR^{C \times T \times I_0 \times \cdots \times I_{N-1}}\).
Then a regular 3D convolution has 
\(C \times T \times \left(\prod_{k=0}^{N-1} I_k\right)\)
parameters. 
By contrast, our HO-CP convolution with rank \(R\) has only 
\(R\left( C + T + \sum_{k=0}^{N-1} I_k\right) + 1\) parameters. The \(+1\) term accounts for the weights \(\myvector{\lambda}\).
For instance, for a 3D convolution with a cubic kernel (of size \(K \times K \times K\), a regular 3D convolution would have \(CTK^3\) parameters, versus only \(R(C + T + 3K)\) for our proposed factorized version.

\begin{figure}
  \centering
  \includegraphics[width=0.8\linewidth]{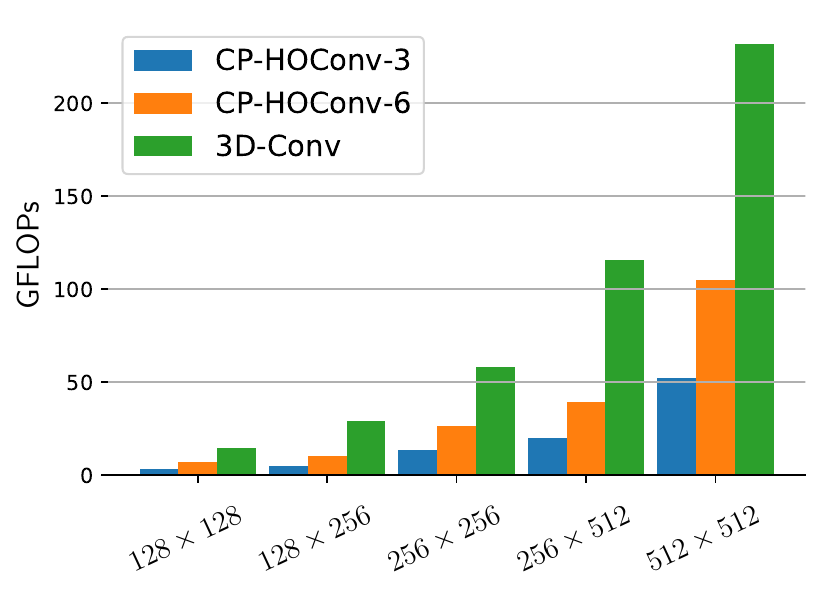}
  \caption{\textbf{Comparison of the number of Giga-FLOPs between regular 3D convolutions and our proposed method.} We consider inputs of size \(32 \times 32 \times 16\), and vary the numer of the input and output channels (the x-axis shows input $\times$ output channels). Our proposed CP-HO convolution, here for a rank equal to 6 and 3 times the input channels (\emph{CP-HOConv-6} and \emph{CP-HOConv-3}), has significantly less FLOPs than regular convolution (\emph{3D-Conv}).}
  \label{fig:FLOP}
\end{figure}

This reduction in the number of parameters translates into much more efficient operation in terms of floating point operations (FLOPs). We show, in Figure~\ref{fig:FLOP}, a visualisation of the number of Giga FLOPs (GFLOPs, with 1GFLOP = \(1e^{9}\) FLOPs), for both a regular 3D convolution and our proposed approach, for an input of size \(32 \times 32 \times 16\), varying the number of input and output channels, with a kernel size of \(3 \times 3 \times 3\).

\paragraph{Higher-Order Transduction}
Here, we introduce \emph{transduction}, which allows to first train an \(N\)--dimensional convolution and expand it to \((N + K)\) dimensions, $K>0$. 

Thanks to the efficient formulation introduced in Equation~\ref{eq:ho-conv-3}, we now have an effective way to go from \(N\) dimensions to \(N+1\). We place ourselves in the same setting as for Equation~\ref{eq:ho-conv-3}, where we have a regular N-D convolution with spatial dimensions and extend the model to \(N+1\) dimensions. To do so, we introduce a new factor \(\mymatrix{U}^{(K_{N+1})} \in \myR^{(K^{N+1} \times R)}\) corresponding to the new \(N+1\myth\) dimension. 
The final formulation is then :
\begin{equation}
    \label{eq:ho-conv-4}
    \mytensor{F} = 
        \hat \rho\left(\Psi
        \left(\mytensor{X} \times_0 \mymatrix{U}^{(T)} \right)
        \right)
    \times_0 \left( \mydiag(\myvector{\lambda}) \mymatrix{U}^{(C)} \right),
\end{equation} with  \(\hat \rho(\mytensor{X}) = \rho(\mytensor{X}) \myconv_{N+1} \mymatrix{U}^{(K_{N+1})}\).

Note that only the new factor needs to be trained, e.g. transduction can be done by simply training \(K_{N+1} \times R\) additional parameters.

\paragraph{Automatic Rank Selection}
Our proposed factorized higher-order convolutions introduce a new additional parameter, corresponding to the rank of the factorization. This can be efficiently incorporated in the formulation by introducing a vector of weights, represented by \(\myvector{\lambda}\) in Equation~\ref{eq:ho-conv-4}. This allows us to automatically tune the rank of each of the layers by introducing an additional \emph{Lasso} term in the loss function, e.g., an \(\ell_1\) regularization on \(\myvector{\lambda}\). Let \(\myvector{\lambda}_l\) be the vector of weights for each layer \(l \in \myrange{0}{L-1}\) of our neural network, associated with a loss \(\mathcal{L}\). The overall loss with regularization  will become \(\mathcal{L}_{\text{reg}} = \mathcal{L} + \gamma \sum_{l=0}^{L-1} |\myvector{\lambda}_l|\), where \(\gamma\) controls the amount of sparsity in the weights.

\section{Experimental setting}\label{sec:experiments}

\paragraph{Datasets for In-The-Wild Affect Estimation}
We validate the performance of our approach on established large scale datasets for continuous affect estimation in-the-wild.

\textbf{AffectNet~\cite{Mollahosseini2017}} is a large static dataset of human faces and labelled in terms of facial landmarks, emotion categories as well as valence and arousal values. It contains more than a million images, including \(450,000\) manually labelled by twelve expert annotators. 

\textbf{AFEW-VA~\cite{kossaifi2017afew}} is composed of video clips taken from feature films and accurately annotated \emph{per frame} in terms of continuous levels of valence and arousal. Besides \(68\) accurate facial landmarks are also provided for each frame.

\textbf{SEWA~\cite{kossaifi2019sewa}} is the largest video dataset for affect estimation in-the-wild. It contains over \(2000\) minutes of audio and video data annotated in terms of facial landmarks, valence and arousal values. 
It contains 398 subjects from six different cultures, is gender balanced and uniformly spans the age range 18 to 65.

\paragraph{Implementation details}
We implemented all models using PyTorch~\cite{pytorch} and TensorLy~\cite{tensorly}. In all cases, we divided the dataset in subject independent training, validation and testing sets. 
For our factorized higher-order convolutional network, we further removed the flattening and fully-connected layers and replace them with a single tensor regression layer~\cite{kossaifi2018tensor} in order to fully preserve the structure in the activations.
For training, we employed an Adam optimizer \cite{kingma2014adam} and validated the learning rate in the range $[10^{-5};0.01]$, the beta parameters in the range $[0.0;0.999]$ and the weight decay in the range $[0.0;0.01]$ using a randomized grid search. We also decreased the learning rate by a factor of $10$ every $15$ epochs. The regularization parameter $\gamma$ was validated in the range $[10^{-4};1.0]$ on AffectNet and consequently set to $0.01$ for all other experiments.

For our baseline we use both a 3D ResNet and a ResNet 2+1D~\cite{tran2018closer}, both with a ResNet-18 backbone. For our method, we use the same ResNet-18 architecture but replace each of the convolutional layers with our proposed higher-order factorized convolution. We initialised the rank so that the number of parameters would be the same as the original convolution. 
When performing transduction, the added temporal factors to the CP convolutions are initialized to a constant value of one. In a first step, these factors are optimized while keeping the remaining parameters fixed. The whole network is then fine-tuned. This avoids the transducted factors to pollute what has already been learnt in the static case. The full process is summarized in Figure~\ref{fig:summary}.

\paragraph{Performance metrics and loss function}
In all cases, we report performance for the \emph{RMSE}, \emph{SAGR}, \emph{PCC}, and \emph{CCC} which are common metrics employed in affect estimation. Let \(\myvector{y}\) be a ground-truth signal and \(\myvector{\hat y}\) the associated prediction by the model. 

The \textbf{RMSE} is the well known Root Mean Square Error: 
$
\text{RMSE}(\myvector{y}, \hat{\myvector{y}}) = \sqrt{\mathbb{E}((\myvector{y}-\hat{\myvector{y}})^{2})}.
$

The \textbf{SAGR} assesses whether the sign of the two signals agree: 
$
SAGR(Y, \hat{\myvector{y}}) = \frac{1}{n}\sum^{n}_{i=1}\delta(\text{sign}(\myvector{y}_{i}), \text{sign}(\hat{\myvector{y}}_{i})).
$

The \textbf{PCC} is the Pearson product-moment correlation coefficient and measures how correlated the two signals are:
$
PCC(\myvector{y}, \hat{\myvector{y}}) = \frac{\mathbb{E}(\myvector{y}-\mu_{\myvector{y}})(\hat{\myvector{y}}-\mu_{\hat{\myvector{y}}})}{\sigma_{\myvector{y}}\sigma_{\hat{\myvector{y}}}}.
$

The \textbf{CCC} is Lin's Concordance Correlation Coefficient and assesses the correlation of the two signals but also how close the two signals are:
$
CCC(\myvector{y}, \hat{\myvector{y}}) = \frac{2\sigma_{\myvector{y}}\sigma_{\hat{\myvector{y}}}\text{PCC}(\myvector{y}, \hat{\myvector{y}})}{\sigma^{2}_{\myvector{y}}+\sigma^{2}_{\hat{\myvector{y}}} + (\mu_{\myvector{y}}-\mu_{\hat{\myvector{y}}})^{2}}.
$

The goal for continuous affect estimation is typically to maximize the correlations coefficients PCC and CCC. However minimizing the RMSE also helps maximizing the correlations as it gives a lower error in each individual prediction. Our regression loss function reflects this by incorporating three terms: 
$\mathcal{L} = \frac{1}{\alpha+\beta+\gamma}(\alpha\mathcal{L}_{RMSE}+\beta\mathcal{L}_{PCC}+\gamma\mathcal{L}_{CCC})
$, with 
$
\mathcal{L}_{RMSE} = \text{RMSE}_{\text{valence}} + \text{RMSE}_{\text{arousal}}
$,

$
\mathcal{L}_{PCC} = 1-\frac{\text{PCC}_{\text{valence}}+\text{PCC}_{\text{arousal}}}{2}
$ 
and 
$
\mathcal{L}_{CCC} = 1-\frac{\text{CCC}_{\text{valence}}+\text{CCC}_{\text{arousal}}}{2}.
$
The coefficients $\alpha$, $\beta$ and $\gamma$ are shake-shake regularization coefficients~\cite{gastaldi2017shake} sampled randomly in the range $[0;1]$ following a uniform distribution. These ensures none of the terms are ignored during optimization. On AffectNet, where discrete classes of emotions are available, we   jointly perform a regression of the valence and arousal values and a classification of the emotional class by adding a cross entropy to the loss function.

\section{Performance evaluation}
In this section, we report the performance of our methods for facial affect estimation in the wild. First, we report results on static images. We then show how the higher-order transduction allows us to extend these models to the temporal domain. In all cases, we compare with the state-of-the-art and show superior results.

\begin{table*}[!h]
    \centering
    \caption{\textbf{Results on the AffectNet dataset}}\label{affectnet}
    \vspace{-8pt}
    \resizebox{0.9\linewidth}{!}{
    \begin{tabular}{l*{9}{l}}
    \toprule
    \multicolumn{2}{l}{ } & \multicolumn{4}{ c }{\bf Valence} & \multicolumn{4}{ c }{\bf Arousal} \\
    \cmidrule{3-6} \cmidrule{7-10}
    \bf Network          & \bf Acc. & \bf RMSE   & \bf SAGR & \bf PCC    & \bf CCC        & \bf RMSE    & \bf SAGR   & \bf PCC       & \bf CCC    \\
    \midrule
    AffectNet baseline~\cite{Mollahosseini2017} & 0.58 & 0.37 & 0.74 & 0.66 & 0.60 & 0.41 & 0.65 & 0.54 & 0.34 \\
    \midrule
    Face-SSD~\cite{jang2019registration} & - & 0.44 & 0.73 & 0.58 & 0.57 & 0.39 & 0.71 & 0.50 & 0.47 \\
    \midrule
    VGG-Face+2M imgs~\cite{kollias2018Generating} & 0.60 & 0.37 & 0.78 & 0.66 & 0.62 & 0.39 & 0.75 & 0.55 & 0.54 \\
    \midrule
    Baseline ResNet-18  & 0.55 & 0.35 & 0.79 & 0.68 & 0.66 & 0.33 & 0.8 & 0.58 & 0.57 \\
    \midrule
    \textbf{Ours} & 0.59 & \bf 0.35 & \bf 0.79 & \bf 0.71 & \bf 0.71 & \bf 0.32 & \bf 0.8 & \bf 0.63 & \bf 0.63\\
    \bottomrule                        
    \end{tabular}
    }
\end{table*} 
\begin{table*}[!h]
    \centering
    \caption{\textbf{Results on the SEWA database}}\label{sewa}
    \vspace{-8pt}
    \resizebox{0.9\linewidth}{!}{
    \begin{tabular}{ll*{8}{c}}
    \toprule
    & \multicolumn{1}{l}{ } & \multicolumn{4}{ c }{\bf Valence} & \multicolumn{4}{ c }{\bf Arousal} \\
    \bf Case & \bf Network          &  \bf RMSE   & \bf SAGR & \bf PCC    & \bf CCC        & \bf RMSE    & \bf SAGR   & \bf PCC       & \bf CCC    \\
    \midrule
    \parbox[t]{5mm}{
            \multirow{4}{*}{\rotatebox[origin=c]{90}{\textbf{Static}}}
    }
    & \cite{kossaifi2019sewa} & - & - & 0.32 & 0.31 & - & - & 0.18 & 0.20 \\
    \cline{2-10}
    & VGG16+TRL~\cite{mitenkova:fg2019} & 0.33 & - & 0.50 & 0.47 & 0.39 & - & 0.44 & 0.39 \\
    \cline{2-10}
    & ResNet-18 & 0.37 & 0.62 & 0.33 & 0.29 & 0.52 & 0.62 & 0.28 & 0.19 \\
    \cline{2-10}
    & \textbf{Ours} &  \bf 0.33 & \bf 0.65 & \bf 0.64 & \bf 0.6 & \bf 0.39 & \bf 0.75 & \bf 0.48 & \bf 0.44\\
    \toprule
    \toprule
    \parbox[t]{5mm}{
            \multirow{4}{*}{\rotatebox[origin=c]{90}{\textbf{Temporal}}}
    } 
    & ResNet-3D & 0.37 & 0.59 & 0.47 & 0.41 & 0.41 & 0.69 & 0.29 & 0.21 \\
    
   \cline{2-10}
    & ResNet-(2+1)D & 0.35 & 0.63 & 0.59 & 0.49 & 0.41 & 0.63 & 0.39 & 0.31 \\
    \cline{2-10}
    & \textbf{Ours}  -- scratch & 0.33 & 0.63 & 0.62 & 0.54 & 0.40 & 0.72 & 0.42 & 0.32 \\
    \cline{2-10}
    
    & \textbf{Ours}  -- transduction & \textbf{0.24} & \textbf{0.69} & \textbf{0.84} & \textbf{0.75} & \textbf{0.32} & \textbf{0.80} & \textbf{0.60} & \textbf{0.52} \\
    \bottomrule                       
\end{tabular}
}
\end{table*}

\begin{table*}[!h]
    \centering
    \caption{\textbf{Results on the AFEW-VA database}}\label{afewva}
    \vspace{-8pt}
    \resizebox{0.9\linewidth}{!}{
    \begin{tabular}{ll*{8}{c}}
    \toprule
    & \multicolumn{1}{l}{ } & \multicolumn{4}{ c }{\bf Valence} & \multicolumn{4}{ c }{\bf Arousal} \\
    \bf Case & \bf Network          &  \bf RMSE   & \bf SAGR & \bf PCC    & \bf CCC        & \bf RMSE    & \bf SAGR   & \bf PCC       & \bf CCC    \\
    \midrule
    \parbox[t]{5mm}{
            \multirow{4}{*}{\rotatebox[origin=c]{90}{\textbf{Static}}}
    }
    &  RF Hybrid DCT~\cite{kossaifi2017afew} & 0.27 & - & 0.407 & - & 0.23 & - & 0.45 & - \\

    \cline{2-10}
    & ResNet50+TRL \cite{mitenkova:fg2019} & 0.40 & - & 0.33 & 0.33 & 0.41 & - & 0.42 & 0.4 \\
    \cline{2-10}
    
    & ResNet-18  & 0.43 & 0.42 & 0.05 & 0.03 & 0.41 & 0.68 & 0.06 & 0.05 \\
    \cline{2-10}
    \cline{2-10}
    
    & \textbf{Ours} & \bf 0.24 & \bf 0.64 & \bf 0.55 & \bf 0.55 & \bf 0.24 & \bf 0.77 & \bf 0.57 & \bf 0.52 \\

    \toprule
    \toprule
    \parbox[t]{5mm}{
            \multirow{5}{*}{\rotatebox[origin=c]{90}{\textbf{Temporal}}}
    } 
    & 
    Baseline ResNet-18-3D & 0.26 & 0.56 & 0.19 & 0.17 & 0.22 & 0.77 & 0.33 & 0.29 \\
    
    \cline{2-10}
    & ResNet-18-(2+1)D & 0.31 & 0.50 & 0.17 & 0.16 & 0.29 & 0.73 & 0.33 & 0.20 \\
    
    \cline{2-10}
    & AffWildNet \cite{kollias2019deep} & - & - & 0.51 & 0.52 & - & - & 0.575 & 0.556 \\
    
    \cline{2-10}
    & \textbf{Ours} -- scratch& 0.28 & 0.53 & 0.12 & 0.11 & 0.19 & 0.75 & 0.23 & 0.15 \\
    
    \cline{2-10}
    \cline{2-10}
    
    & \textbf{Ours} -- transduction  &  \textbf{0.20} & \textbf{0.67} & \textbf{0.64} & \textbf{0.57} & \textbf{0.21} & \textbf{0.79} & \textbf{0.62} & \textbf{0.56} \\

    \bottomrule                        
\end{tabular}
}
\end{table*}
 
\paragraph{Static affect analysis in-the-wild with factorized CNNs} 
First we show the performance of our models trained and tested on individual (static) imagery. We train our method on AffectNet, which is the largest database but consists of static images only. There, our method outperforms all other works by a large margin (Table~\ref{affectnet}). We observe similar results on SEWA (Table~\ref{sewa}) and AFEW-VA (Table~\ref{afewva}). In the supplementary document, we also report results on LSEMSW~\cite{hu2018deep} and CIFAR10~\cite{krizhevsky2009learning}.

\paragraph{Temporal prediction via higher-order transduction}
We then apply transduction as described in the method section to convert the static model from SEWA (Table~\ref{sewa}, temporal case) and AFEW-VA (Table~\ref{afewva}, temporal case) to the temporal domain, where we simply train the added temporal factors. This approach allows to efficiently train temporal models on even small datasets. Our method outperforms other approaches, despite having only $11$ million parameters, compared to $33$ million for the corresponding 3D ResNet18, and $31$ million parameters for a (2+1)D ResNet-18.
Interestingly, in all cases, we notice that valence is better predicted than arousal, which, is in line with finding by psychologists that humans are better at estimating valence from visual data~\cite{russel2003,Kroschel07}.

Using the automatic rank selection procedure detailed in section~\ref{sec:ho-cp-conv}, we let the model learn end-to-end the rank of each of the factorized higher-order convolutions.
We found that on average, 8 to 15\% of the parameters can be set to zero for optimal performance. In practice, about 1 million (of the 11 million) parameters were set to zero by the Lasso regularization. An in-depth study of the effect of the automatic rank selection is provided in the supplementary document.

\section{Conclusion}
\vspace{-4pt}
We established the link between tensor factorizations and efficient convolutions, in a unified framework. 
Based on this, we proposed a factorized higher-order (N-dimensional) convolutional block.
This results in efficient models that outperform traditional networks, while being more computationally and memory efficient.
We also introduced a higher-order transduction algorithm for converting the expressive power of trained \(N\)--dimensional convolutions to any \(N+K\) dimensions. We then applied our approach to continuous facial affect estimation in naturalistic conditions. Using transduction, we transferred the performance of the models trained on static images to the temporal domain and reported state-of-the-art results in both cases.

{\small
\bibliographystyle{ieee_fullname}

}

\clearpage

\section*{APPENDIX}
Here, we give additional details on the method, as well as results on other tasks (image classification on CIFAR-10, and gesture estimation from videos).

\section*{Dimensional model of affect}

\begin{figure}[h!]
  \centering
  \includegraphics[width=5cm, page=1]{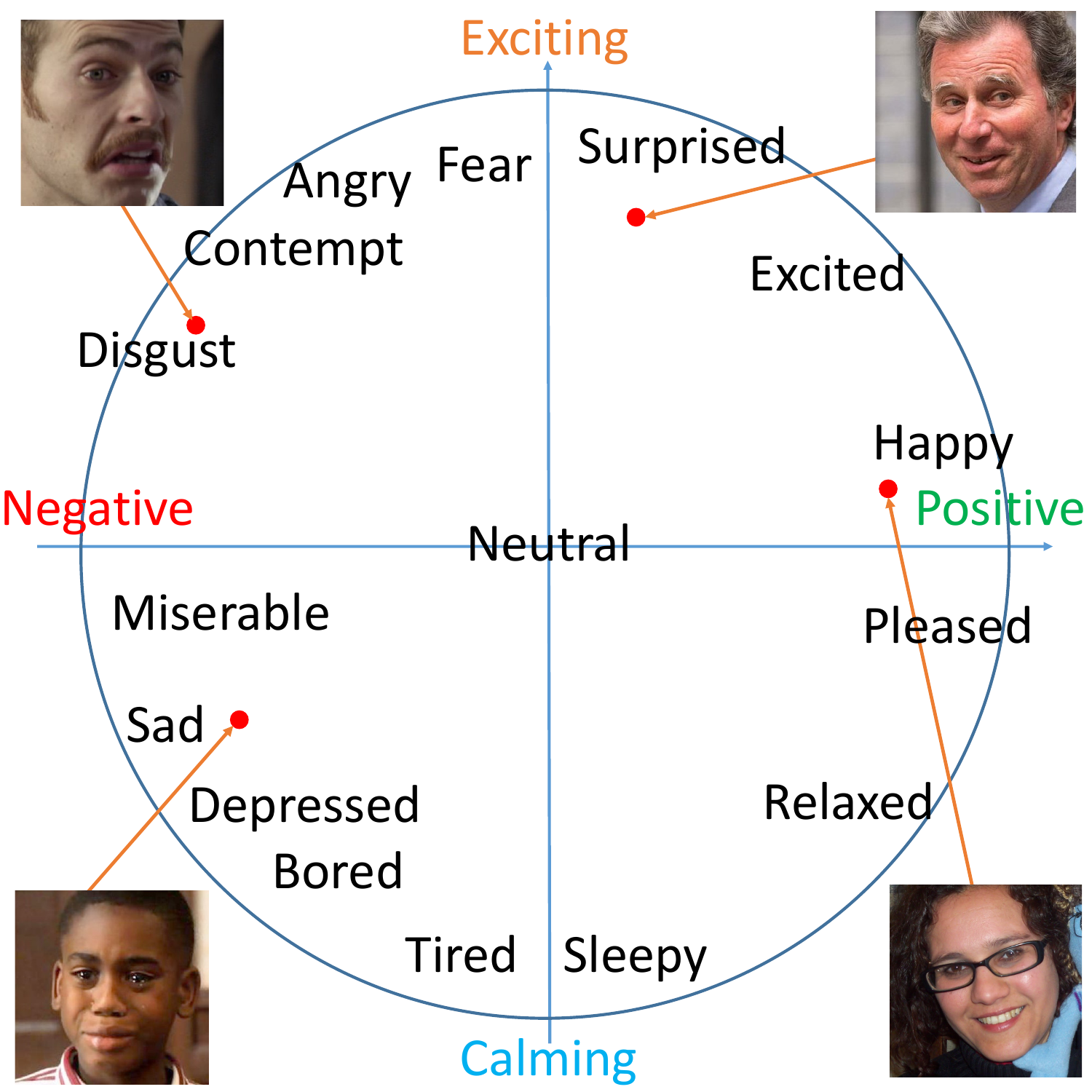}
  \caption{The valence and arousal circumplex. This dimensional model of affect covers the continuous range of emotions displayed by human on a daily basis. The images are taken from the AffectNet dataset~\cite{Mollahosseini2017}}
  \label{fig:va_circle}
\end{figure}

Discrete emotional classes are too coarse to summarize the full range of emotions displayed by humans on a daily basis. This is the reason why finer, dimensional affect models, such as the valence and arousal are now favoured by psychologists~\cite{russel2003}.In this circumplex, which can be seen in Figure~\ref{fig:va_circle}, the valence level corresponds to how positive or negative an emotion is, while the arousal level explains how calming or exciting the emotion is. 

\begin{figure*}[th]
    \centering
    \includegraphics[width=1\textwidth]{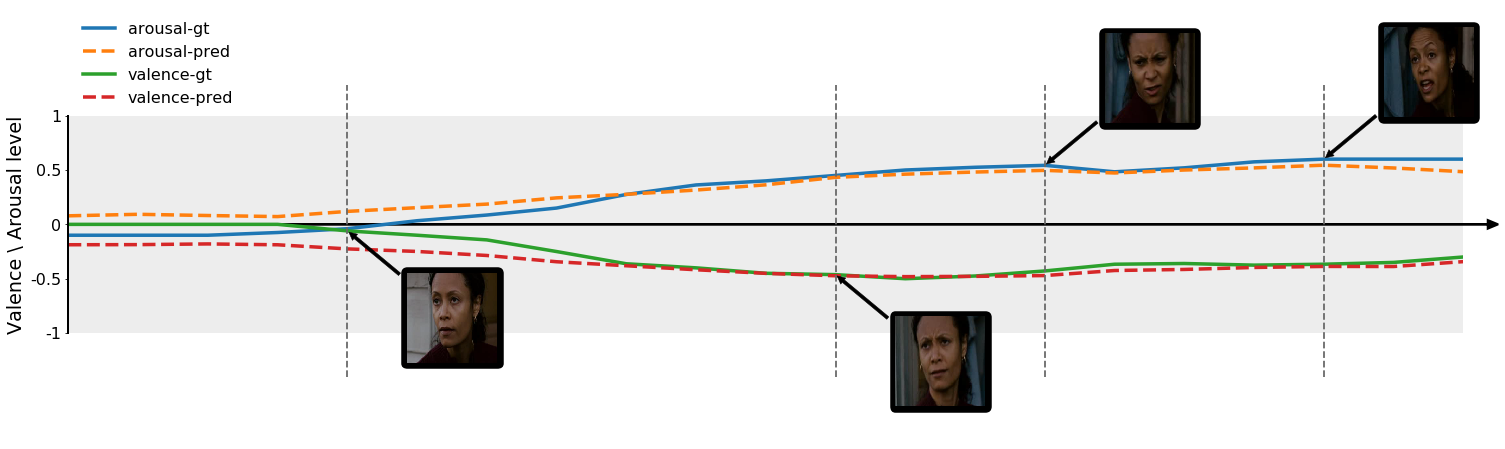} \vspace{-10pt}
    \caption{\textbf{Evolution of the ground-truth (\emph{gt}) and predicted (\emph{pred}) levels of valence and arousal} as a function of time, for one of the test videos of the \emph{AFEW-VA} dataset.}
    \label{fig:my_label}
\end{figure*}

 A visualization of the prediction of valence and arousal of our model can be seen in Figure~\ref{fig:my_label}, along with some representative frames. 

\section*{Automatic rank selection}

Using the automatic rank selection procedure detailed in the method section, we let the model learn end-to-end the rank of each of the factorized higher-order convolutions.

\begin{figure}[h!]
    \centering
    \includegraphics[width=1\linewidth]{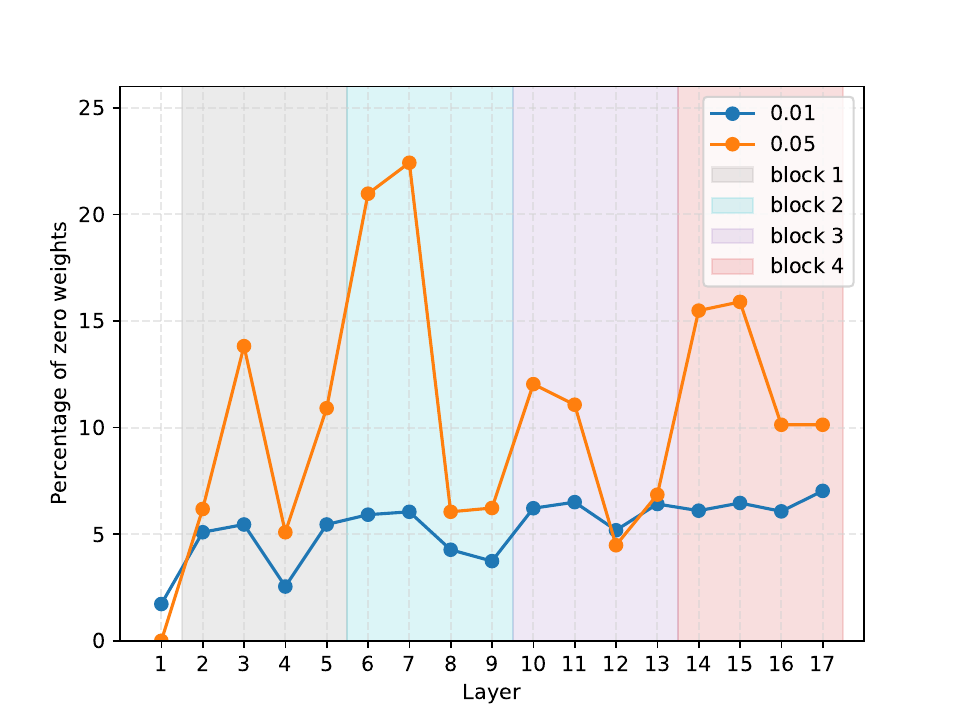}
    \vspace{-25pt}
    \caption{Sparsity induced by the automatic rank selection at each layer of the network (ResNet-18 backbone).}
    \label{fig:sparsity}
    \vspace{-10pt}
\end{figure}

In Figure~\ref{fig:sparsity}, we show the number of parameters set to zero by the network for a regularization parameter of $0.01$ and $0.05$. The lasso regularization is an easy way to automatically tune the rank. We found that on average 8 to 15\% of the parameters can be set to zero for optimal performance. In practice, about 1 million parameters were removed thanks to this regularization.

In Table \ref{tab:parameters} we report the number of parameters of spatio-temporal baselines and compare it to our CP factorized model. Besides having less parameters, our approach has the advantage of having a very low number of temporal parameters which facilitates the training on spatio-temporal data once it has been pretrained on static data.

\begin{table}[h]
  \caption{\textbf{Number of parameters optimized to train the temporal model}}
  \label{tab:parameters}
  \centering
  \resizebox{1\linewidth}{!}{
  \begin{tabular}{cccc}
    \toprule
    \textbf{Network}  & \textbf{\thead{Total \\ \# parameters}}  & \textbf{\thead{\# parameters \\ removed \\ with LASSO}} & \textbf{\thead{\# parameters \\ optimized \\ for video}}\\
    \midrule
    ResNet18-(2+1)D & 31M & - & 31M\\
    \midrule
    ResNet-18-3D & 33M & - & 33M \\
    \midrule
    \textbf{Ours} $[\lambda = 0.01]$ & 11M & 0.7 & 0.24 \\
    \midrule
    \textbf{Ours} $[\lambda = 0.05]$ & 11M & 1.3M & 0.24 \\
    \bottomrule
  \end{tabular}
  }
\end{table}

\section*{Results on LSEMSW}

\textbf{LSEMSW~\cite{hu2018deep}} is the Large-scale Subtle Emotions and Mental States in the Wild database. It contains more than \(175,000\) static images annotated in terms of subtle emotions and cognitive states such as helpless,  suspicious, arrogant, etc. We report results on LSEMSW in table.~\ref{LSEMSW}.

\begin{table}[h]
    \centering
    \small
    \caption{Results on the LSEMSW database}\label{LSEMSW}\vspace{-5pt}
    \begin{tabular}{lc}
    \toprule
    \textbf{Method} & \textbf{Accuracy}\\
    \midrule
    ResNet 34~\cite{hu2018deep} & 28.39 \%\\
    \textbf{Ours} & \textbf{34.55\%}\\
    \bottomrule
    \normalsize
    \vspace{-20pt}
    \end{tabular}
\end{table}

\section*{Results on CIFAR-10}
While in the paper we focus on affect estimation, we report here results on a traditional image classification dataset, CIFAR 10.

\textbf{CIFAR-10~\cite{krizhevsky2009learning}} is a dataset for image classification composed of \(10\) classes with \(6,000\) images which, divided into \(5000\) images per class for training and \(1000\) images per class for testing, on which we report the results.

We used a MobileNet-v2 as our baseline. For our approach, we simply replaced the full MobileNet-v2 blocks with ours (which, in the 2D case, differs from MobileNet-v2 by the use of two separable convolutions along the spatial dimensions instead of a single 2D kernel).
We kept all the parameters the same for all experiments to allow for fair comparison and reported performance averaged across \(3\) runs. The standard deviation was \(0.033\) for MobileNet-v2 and \(0.036\) for our approach. We optimized the loss using stochastic gradient descent with a mini-batch size of \(128\), starting with a learning rate of \(0.1\), decreased by a factor of \(10\) after \(150\) and \(250\) epochs, for a total of \(400\) epochs, with a momentum of \(0.9\). For MobileNet-v2, we used a weight decay of \(4e^{-5}\) and \(1e^{-4}\) for our approach. Optimization was done on a single NVIDIA 1080Ti GPU.

\begin{table}[bt]
  \caption{\textbf{Results on the CIFAR-10 dataset}}
  \label{tab:cifar}
  \centering
  \begin{tabular}{llc}
    \toprule
    \textbf{Network}  & \textbf{\# parameters} & \textbf{Accuracy (\%)} \\
    \midrule
    MobileNet-v2             & $2.30$M  & $94$   \\ 
    \textbf{Ours}  & $2.29$M  & $94$  \\ 
    \bottomrule
  \end{tabular}
\end{table}

We compared our method with a MobileNet-v2 with a comparable number of parameters, Table~\ref{tab:cifar}. Unsurprisingly, both approach yield similar results since, in the 2D case, the two networks architectures are similar. It is worth noting that our method has marginally less parameters than MobileNet-v2, for the same number of channels, even though that network is already optimized for efficiency.

\section*{Results on gesture estimation}
In this section, we report additional results on the Jester dataset. We compare a network that employs regular convolutional blocks to the same network that uses our proposed higher-order factorized convolutional blocks.

\begin{figure*}[ht]
  \centering
  \includegraphics[width=1\textwidth,trim={5 5 5 5},clip]{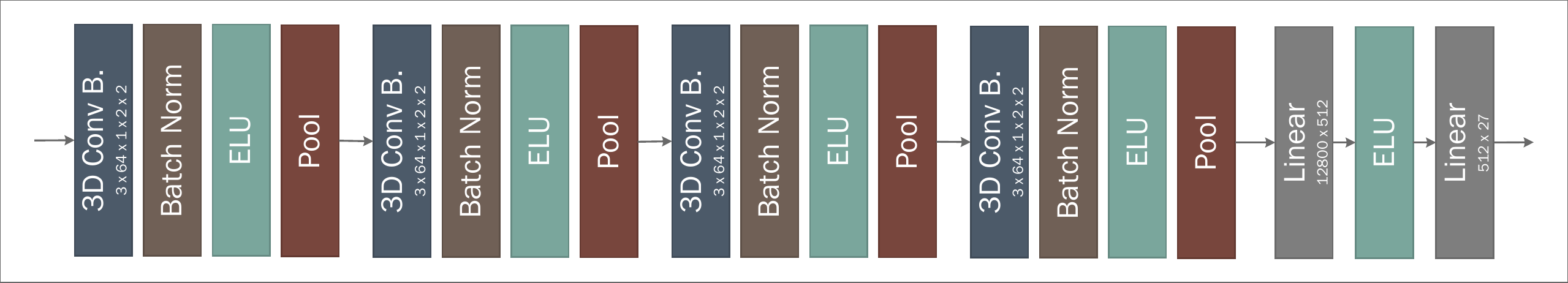}
  \caption{\textbf{Architecture of our 3D convolutional network}. We employed the same architecture for both our baseline and our approach, where the only difference is the 3D convolutional block used (3D Conv B): for the baseline a regular 3D conv, and for our method, our proposed HO-CP conv-S. Each convolution is followed by a batch-normalisation, non-linearity (ELU) and a max pooling (over \(2\times2\times2\) non-overlapping regions).}
  \label{fig:3dconv-arch}
\end{figure*}

\textbf{20BN-Jester v1} is a dataset\footnote{Dataset available at \url{https://www.twentybn.com/datasets/jester/v1}.} composed of \(148,092\) videos, each representing one of \(27\) hand gestures (e.g. \emph{swiping left}, \emph{thumb up}, etc). Each video contains a person performing one of gestures in front of a web camera. Out of the $148,092$ videos $118,562$ are used for training, $14,787$ for validation on which we report the results. 

For the 20BN-Jester dataset, we used a convolutional column composed of \(4\) convolutional blocks with kernel size \(3 \times 3 \times 3\), with respective input and output of channels: $(3, 64), (64, 128), (128, 256)$ and $(256, 256)$, followed by two fully-connected layers to reduce the dimensionality to \(512\) first, and finally to the number of classes. Between each convolution we added a batch-normalisation layer, non-linearity (ELU) and \(2\times 2 \times 2\) max-pooling. The full architecture is graphically represented in Figure~\ref{fig:3dconv-arch} of the architecture detailed in the paper, for clarity. 

For our approach, we used the same setting but replaced the 3D convolutions with our proposed block and used, for each layer, \(6 \times n_\text{input-channels}\) for the rank of the HO-CP convolution. The dataset was processed by batches of \(32\) sequences of RGB images, with a temporal resolution of \(18\) frames and a size of \(84 \times 84\). The loss is optimized by mini-batches of \(32\) samples using stochastic gradient descent, with a starting learning-rate of \(0.001\), decreased by a factor of \(10\) on plateau, a weight decay of \(1e^{-5}\) and momentum of \(0.9\). All optimization was done on \(2\) NVIDIA 1080Ti GPUs.

\begin{table}[htb]
  \caption{\textbf{Results on the 20BN-Jester Dataset}}
  \label{tab:jester}
  \centering
  \resizebox{1\linewidth}{!}{
  \begin{tabular}{llll}
    \toprule
    &  \multicolumn{1}{c}{\textbf{\#conv}} & \multicolumn{2}{c}{\textbf{Accuracy (\%)}}                   \\
    \cmidrule(r){3-4}
    \multirow{-2}{*}{\textbf{Network}}  & \textbf{parameters} & \textbf{Top-1}     & \textbf{Top-5} \\
    \midrule
    3D-ConvNet                       & $2.9$M  & $83.2$            &  $97.0$  \\
    HO-CP ConvNet (\textbf{Ours})    & $1.2$M  & $83.8$            &  $97.4$   \\
    HO-CP ConvNet-S (\textbf{Ours})  & $1.2$M  & $\mathbf{85.4}$   &  $\mathbf{98.6}$ \\ 
    \bottomrule
  \end{tabular}
  }
\end{table}

\textbf{Results for 3D convolutional networks}
For the 3D case, we test our Higher-Order CP convolution with a regular 3D convolution in a simple neural network architecture, in the same setting, in order to be able to compare them. Our approach is more computationally efficient and gets better performance as shown in Table~\ref{tab:jester}. In particular, the basic version without skip connection and with RELU (emph{HO-CP ConvNet}) has \(1.7\) million less parameters in the convolutional layers compared to the regular 3D network, and yet, converges to better Top-1 and Top-5 accuracy. The version with skip-connection and PReLU (\emph{HO-CP ConvNet-S}) beats all approaches.

\section*{Algorithm for our CP convolutions}

We summarize our efficient higher-order factorized convolution in algorithm~\ref{alg:ho-cp-conv}.

\begin{algorithm}[t!] 
\caption{Higher-order CP convolutions} 
\label{alg:ho-cp-conv} 
\begin{algorithmic} 
    \STATE \textbf{Input}\\
            \hspace*{\algorithmicindent} \(\bullet\) Input activation tensor $\mytensor{X}\in \myR^{C \times D_0 \times \cdots \times D_N}$\\
            \hspace*{\algorithmicindent}  \(\bullet\) CP kernel weight tensor $\mytensor{W}$:\\
            \hspace*{\algorithmicindent} \qquad $\mytensor{W} = \mykruskal{\mymatrix{U}^{(T)}, \mymatrix{U}^{(C)}, \mymatrix{U}^{(K_0)}, \cdots, \mymatrix{U}^{(K_N)}}$\\
            \hspace*{\algorithmicindent}  \(\bullet\) Skip connection weight matrix $ \mymatrix{U}^{(S)} \in \myR^{T \times C}$\\
    \textbf{Output}	\hspace*{\algorithmicindent} Efficient factorized N-D convolution \(\mytensor{X} \myconv \mytensor{W}\)
    \STATE $\mytensor{H} \Leftarrow \mytensor{X} \times_0 \mymatrix{U}^{(C)}$
    \FOR{ i:=1 \TO \(N - 2\)}
        \STATE $\mytensor{H} \Leftarrow \mytensor{H} \myconv_i \mymatrix{U}^{(K_i)}$ (\(1\)--D conv along the i\myth mode)
        \STATE $\mytensor{H} \Leftarrow \text{PReLU}\left(\mytensor{H}\right) $ or $\text{ReLU}\left(\mytensor{H}\right)$ \emph{[optional]}
        \STATE $\mytensor{H} \Leftarrow \text{Batch-Norm}\left(\mytensor{H}\right)$ \emph{[optional]}
    \ENDFOR
    \STATE $\mytensor{H} \Leftarrow \mytensor{H} \times_1 \mymatrix{U}^{(T)}$
    \IF{ skip-connection} 
        \RETURN $\mytensor{H} + \mytensor{X} \times_0 \mymatrix{U}^{(S)} $
    \ELSE        
        \RETURN $\mytensor{H}$
    \ENDIF
\end{algorithmic}
\end{algorithm}

\end{document}